\PassOptionsToPackage{table}{xcolor}
\documentclass{article}

\usepackage[preprint]{neurips_2026}

\usepackage[utf8]{inputenc} 
\usepackage[T1]{fontenc}    
\usepackage{hyperref}       
\usepackage{url}            
\usepackage{booktabs}       
\usepackage{amsfonts}       
\usepackage{nicefrac}       
\usepackage{microtype}      
\usepackage{xcolor}         
\newcommand{\ourmeth}{\textit{GraphThinker}}
\usepackage{graphicx}
\usepackage{wrapfig}
\usepackage{multirow}
\usepackage{caption}
\usepackage{subcaption}
\usepackage{amsmath}
\newcommand{\shaded}[1]{\colorbox{black!6}{#1}}

\title{GraphThinker: Reinforcing Temporally Grounded Video Reasoning with Event Graph Thinking}

\author{Zixu Cheng\textsuperscript{1}, Da Li\textsuperscript{1,2}, Jian Hu\textsuperscript{1}, Yuhang Zang\textsuperscript{3}, Ziquan Liu\textsuperscript{1}, Shaogang Gong\textsuperscript{1}, Wei Li\textsuperscript{4}\thanks{corresponding author}\\
\textsuperscript{1}Queen Mary University of London,
\textsuperscript{2}Samsung AI Centre Cambridge,\\
\textsuperscript{3}Shanghai Artificial Intelligence Laboratory,
\textsuperscript{4}Nanyang Technological University\\
{\tt\small \{zixu.cheng,jian.hu,ziquan.liu,s.gong\}@qmul.ac.uk,wei.l@ntu.edu.sg} } 

\begin{document}

\maketitle

\begin{abstract}
Video reasoning requires a fine-grained understanding of the temporal dependencies and event-level relations between objects and events in videos. Current Multimodal Large Language Models (MLLMs) are prone to severe temporal hallucinations in video reasoning. An underlying cause of these hallucinations is weak visual-temporal grounding and the lack of explicit structure for modelling event relations. Models often rely on auxiliary text, such as dense captions, rather than explicitly anchoring their reasoning in actual visual evidence. However, these textual representations are inherently unstructured and fail to provide explicit causal constraints needed to guide the model's reasoning. In this work, we propose \ourmeth{}, a reinforcement finetuning method that constructs a structured event representation of a video and enforces visual grounding to jointly reduce reasoning hallucinations. 
Specifically, we employ an MLLM to construct an Event-based Video Scene Graph (EVSG) that captures both intra- and inter-event relations, guiding a structured video reasoning process. Moreover, we address the weak grounding issue by introducing a novel visual attention reward during reinforcement finetuning that encourages the model to actively attend to reliable visual cues. On the RexTime dataset, \ourmeth{} achieves an over \textbf{4\%} improvement in IoU ($=0.3$) for moment localisation. On the VidHalluc dataset, \ourmeth{} achieves a \textbf{9.8\%} improvement in reducing temporal sequence hallucination and a \textbf{7.6\%} gain in Binary QA in reducing action hallucination, compared to the state-of-the-art methods.
\end{abstract}

\section{Introduction}
\label{sec:intro}
Video reasoning requires understanding temporal event transitions and relational dependencies across video contexts to answer complex queries~\cite{chen2024rextime,cheng2025v,fang2024mmbench,han2025videoespresso,li2023intentqa,zhang2025towards,fu2025video}. Such capabilities are crucial for high-level applications, including instructional video understanding~\cite{bai2025qwen2,wang2025internvl3,zhang2025videollama}, embodied decision-making~\cite{yang2025thinking,li2024embodied,fung2025embodied}, and assistive AI systems~\cite{gia2025real,verma2025causal}, that rely on reliable video reasoning.

\begin{figure*}[t]
  \centering
  \includegraphics[width=1.0\linewidth]{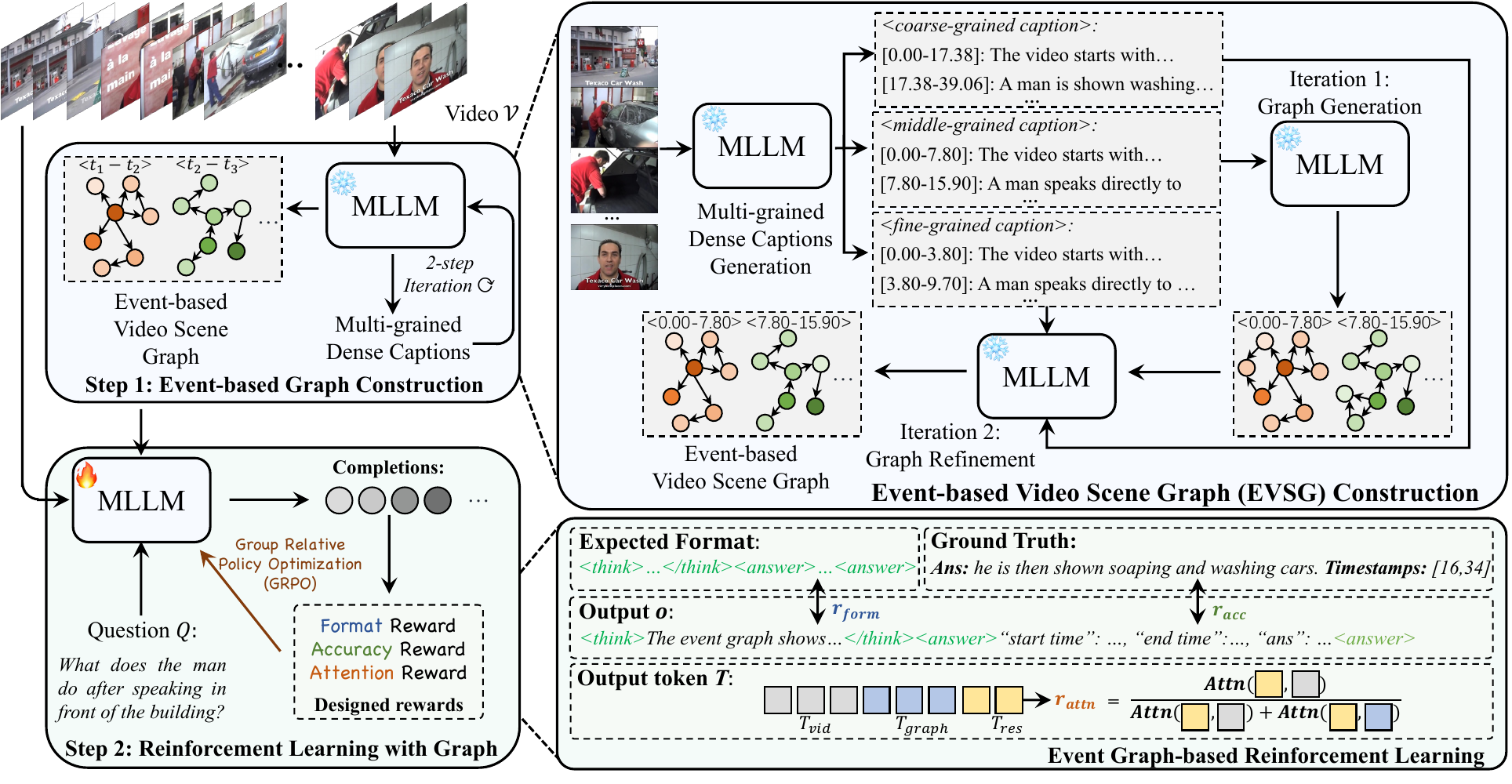}
  \vspace{-17pt}
  \caption{An overview of the \ourmeth{}. It first employs an MLLM to generate multi-grained \textit{dense captions} for a video, and then prompts the same MLLM to select keywords as nodes to construct an event-based graph (EVSG). This EVSG is then used to serve as a fine-grained representation of structured event relations for reasoning. Given EVSGs, we further develop an Event Graph-based RL finetuning method with a visual attention reward to ground more attentive visual evidence. Together, \ourmeth{} achieves visually more grounded and temporally more consistent video reasoning. 
  }
  \vspace{-17pt}
  \label{fig:pipeline}
\end{figure*}

Recent efforts have substantially improved video understanding by leveraging the reasoning of Multimodal Large Language Models (MLLMs)~\cite{zhang2025vtimecot,han2025videoespresso,ghazanfari2025chain,qiu2025step, feng2025video,wang2025time,li2025reinforcement}, which are often further enhanced through post-training, such as Supervised Fine-Tuning (SFT)~\cite{zhang2023video} or Reinforcement Fine-Tuning (RFT)~\cite{feng2025video}. SFT-based approaches~\cite{huang2024lita,guo2024trace,zhang2025video,zhang2025vtimecot,han2025videoespresso,ghazanfari2025chain,qiu2025step} teach MLLMs to follow annotated reasoning paths, while RFT-based methods~\cite{feng2025video,wang2025time,li2025reinforcement,shen2025vlm,chen2025scaling,li2025videochat,hong2025glm,luo2025museg} optimise models with designed reward signals. Despite producing fluent reasoning chains, however, these models remain prone to hallucinations involving event ordering, action sequence, and scene transitions~\cite{liu2025more,luo2025thinking,jian2025look,fan2025sophiavl,chen2025perturbollava}, where reasoning depends on event-temporal structure.

{These hallucinations are caused by two aspects. First, MLLM-based video reasoning exhibits weak visual grounding~\cite{li2025vidhalluc,luo2025dr}, i.e., it tends to rely on linguistic priors rather than on grounding in visual evidence from the video. Current methods, therefore, incorporate auxiliary textual cues, such as dense captions, to provide rich semantic context for video reasoning~\cite{xia2025visionary,li2024temporal,kim2024you,ghazanfari2025chain,han2025videoespresso,qin2025question}. However, these textual representations fail to explicitly encode temporal constraints across events, hindering solid visual reasoning. Second, the MLLM-based video reasoning relies on implicit event-relational modelling. Even when visual context is available, current representations do not explicitly encode the temporal and relational structure between events. Some works incorporate video-level scene graphs to model object relations in a video~\cite{fei2024video,huang2025building,nguyen2025hyperglm}. }
Yet these graphs are often global and coarse, and do not explicitly represent intra- and inter-event relations. Therefore, current representations remain limited in constraining event ordering and cross-event causality, which can confuse MLLMs and induce hallucinations in video reasoning.

To address the above limitations, we propose \ourmeth{}, a reinforcement-finetuning method that constructs \textbf{structured event-level scene graphs} to guide reasoning through explicit event relations. 
Moreover, \ourmeth{} introduces a \textbf{visual-attention reward} during reinforcement finetuning to encourage the model to {attend to evidence-enabled visual cues, thereby mitigating language-only shortcuts and reducing hallucinations}.

Specifically, we construct structured Event-based Video Scene Graphs (EVSGs) that explicitly encode fine-grained event relations to guide the MLLM's reasoning. Our EVSGs are constructed without human annotations by a \emph{self-generate} and \emph{self-refine} process. We first use an MLLM to generate multi-grained event-level dense captions, and then prompt it to refine these captions into structured scene graphs. Each graph captures \textit{intra-event} relations through object nodes and their semantic interactions within an event, while \textit{inter-event} relations are modelled as timestamp-based temporal edges connecting event subgraphs. 
Although EVSG remains textual, it provides structured event-level evidence rather than unstructured caption sequences, organising video content into timestamped events, triplets, and temporal links to reduce ambiguity and support consistent temporal reasoning.

{Moreover, we further enhance the MLLM's temporally grounded reasoning capability with reinforcement-finetuning. We introduce a visual attention reward that encourages the MLLM to actively attend to reliable visual cues during video reasoning, rather than relying solely on the input EVSGs.} Together, the formed textual EVSGs and visual attention reward provide complementary benefits that enable MLLMs to achieve visually grounded and temporally consistent video reasoning.

We summarise our contributions as follows: 
1) We propose \ourmeth{}, a reinforcement finetuning method that constructs an Event-based Video Scene Graph (EVSG) to model intra- and inter-event relations for structured video reasoning. 
{2) \ourmeth{} introduces a novel visual attention reward that encourages visual grounding of EVSG-guided reasoning during reinforcement finetuning, reducing over-reliance on textual graph cues and thereby mitigating reasoning hallucinations. }
3) \ourmeth{} achieves notable improvements over existing state-of-the-art video reasoning MLLMs on two benchmarks, \emph{RexTime} and \emph{VidHalluc}. Moreover, it is more inference-efficient than prior SoTA methods and incurs only negligible additional overhead compared with vanilla GRPO-trained MLLMs, when using the reusable EVSG representation.

\section{Related Works}
\label{sec:related_works}

\noindent \textbf{Video Scene Graph Generation}
Video scene graphs (VSGs) decompose videos into objects and their pairwise relations to model how objects interact over time~\cite{ji2020action,yang2023panoptic,wu2024sportshhi}. These structured representations bridge low-level perception and high-level reasoning, enabling interpretable video understanding. Traditional VSG methods include two-stage approaches~\cite{cong2021spatial,nguyen2024hig,nag2023unbiased}, which first detect objects and then classify relations, and one-stage approaches~\cite{li2018factorizable}, which predict entities and relations simultaneously. However, both are limited by closed vocabularies and external detectors' results. Open-vocabulary VSG methods~\cite{he2022towards,li2024pixels} often use vision-language models to predict relational triplets in unconstrained semantic spaces, thereby enhancing generalisation to unseen concepts; however, they often produce visually ungrounded or temporally inconsistent graphs. Moreover, existing methods~\cite{fei2024video,nguyen2025hyperglm} typically generate coarse video-level graphs that overlook fine-grained temporal boundaries and event dependencies, leading to imprecise temporal understanding and potential hallucinations in video reasoning. To this end, we propose the event-based video scene graph (EVSG), which explicitly encodes object and event relations over time, providing a structured representation enabling MLLMs to perform visually grounded and temporally consistent video reasoning.

\noindent \textbf{Post-training MLLMs for Video Reasoning}
Video reasoning involves identifying relevant visual evidence and inferring event-level temporal and causal relationships to answer natural language queries. Recent MLLMs address this by post-training optimisation, typically through supervised fine-tuning (SFT) or reinforcement fine-tuning (RFT).
SFT-based methods employ chain-of-thought (CoT) datasets to learn reasoning over video inputs. Specifically, they~\cite{huang2024lita,guo2024trace,zhang2025video, zhang2025vtimecot,han2025videoespresso,ghazanfari2025chain,qiu2025step} train models on annotated chain-of-thought paths, promoting structured and interpretable reasoning, but suffer from generalisation limitations in novel scenarios. RFT-based methods~\cite{feng2025video,wang2025time,li2025reinforcement,shen2025vlm, chen2025scaling,li2025videochat,hong2025glm,luo2025museg} instead optimise reasoning trajectories under reward signals. These methods offer greater flexibility yet often over-rely on language priors, resulting in visually ungrounded reasoning~\cite{liu2025more,luo2025thinking,jian2025look,fan2025sophiavl,chen2025perturbollava}. Other methods improve reasoning by visual summaries~\cite{li2024temporal,ghazanfari2025chain,han2025videoespresso}, captions~\cite{xia2025visionary,kim2024you,qin2025question}, or video-level scene graphs~\cite{fei2024video,nguyen2025hyperglm,huang2025building} with additional thinking rewards~\cite{luo2025thinking,fan2025sophiavl, jian2025look}, but they struggle to capture fine-grained event relations, leading to hallucinated reasoning. To address these issues, we develop an event graph-based RFT approach, with a visual attention reward that guides an MLLM to extract the most informative cues in the input video for visually grounded reasoning.

\section{Method}
\label{sec:method}
\subsection{Task Definition}
We apply \ourmeth{} to the task of video reasoning~\cite{chen2024rextime}, a problem closely related to reasoning temporal localization~\cite{huang2024lita} and grounded VQA~\cite{xiao2024can}. The objective is twofold: the model must infer the correct answer to a complex query while simultaneously localising the specific temporal segment that justifies its reasoning. Formally, let $V \in \mathbb{R}^{N \times H \times W \times 3}$ denote an input video with $N$ frames of spatial dimensions $H \times W$, and let $q$ represent a textual question. Our model $\Phi$ jointly predicts a natural language answer $A$ and a supporting temporal window $T = (t^s, t^e)$, where $t^s$ and $t^e$ are the start and end times, respectively. This reasoning task is thus defined as: $[T, A] = \Phi(V, q).$

\begin{figure}[t]
  \centering
  \vspace{-15pt}
  \includegraphics[width=1.0\linewidth]{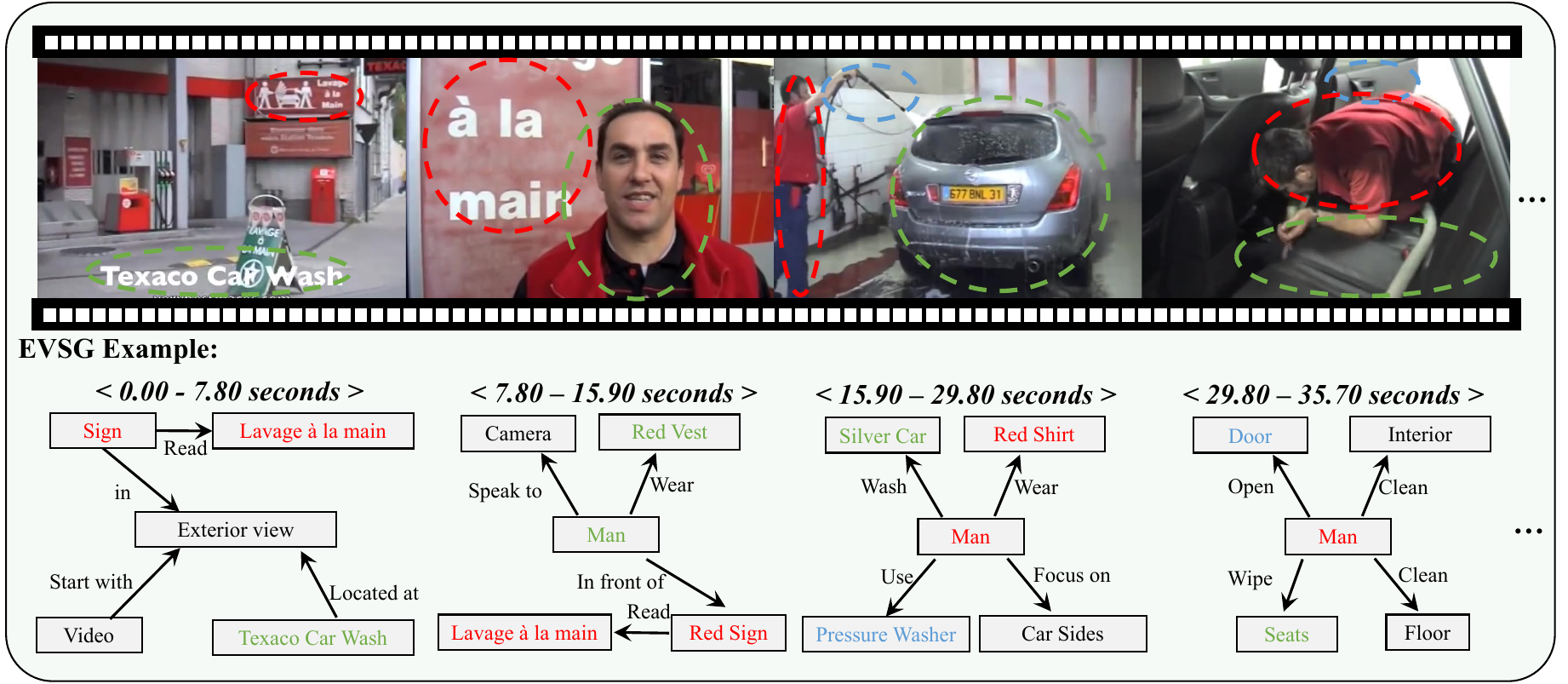}
  \vspace{-15pt}
  \caption{An example of the proposed Event-based Video Scene Graph (EVSG). The EVSG comprises event subgraphs (derived from event-level captions), each subgraph corresponding to start–end timestamps and a set of triplets that represent object interactions and spatial relationships to capture intra-event semantics. Event subgraphs are sequentially linked by timestamp-based edges, forming a hierarchical structure that explicitly models both intra-event and inter-event relations.}
  \vspace{-10pt}
  \label{fig:example}
\end{figure}

\subsection{Event-based Video Scene Graph (EVSG)}
To provide explicit event-level structure for temporally grounded reasoning, we construct an Event-based Video Scene Graph (EVSG). EVSG serves as a structured representation that organises video evidence into event nodes, intra-event entity-relation triplets, and inter-event temporal edges.

Given an input video, we first prompt an MLLM to generate multi-grained dense captions with different event-number constraints, producing coarse, middle, and fine descriptions, denoted as $C^{\text{coarse}}$, $C^{\text{middle}}$, and $C^{\text{fine}}$. 
Multi-grained captioning provides a multi-stage self-consistency mechanism. Coarse captions capture the global event progression, while fine-grained captions preserve local details. During EVSG construction, these descriptions are jointly used to filter inconsistent or weakly supported event descriptions, yielding a more robust structured representation than dense captions.


{Next, we employ the MLLM to construct the EVSG from these self-generated captions via a two-stage iterative process: generation and refinement. Specifically, first the MLLM primarily parses the middle-grained captions $C^{\text{middle}} = \{ c_1, c_2, \dots, c_n \}$ to build an initial event graph, where $n$ is the number of segmented events. Guided by predefined instructions, the MLLM extracts the start and end timestamps for each event description $c_i$ and maps it to a structured event subgraph $e_i$:
\begin{equation}
    e_i = f_{\text{MLLM}}(c_i) = (t_i^s, t_i^e, \mathcal{T}_i),
    \quad 
    \mathcal{T}_i=\{(s,r,u)\},
\end{equation}
where $t_i^s$ and $t_i^e$ denote the start and end timestamps, and $\mathcal{T}_i$ is the set of intra-event $\langle\text{subject--relation--object}\rangle$ triplets. The initial video-level graph is then constructed as:
\begin{equation}
    \mathcal{G}_{\text{init}} = (\mathcal{V}, \mathcal{E}), 
    \quad \text{where} \quad \mathcal{V}=\{e_i\}_{i=1}^{n}, \quad \mathcal{E}=\{(e_i,e_j)\mid t_i^e \leq t_j^s\},
\end{equation}
where each edges $\mathcal{E}$ explicitly encode global temporal dependencies, connecting event subgraph $e_i$ to $e_j$ only if $e_i$ temporally precedes $e_j$ as $t_i^e \le t_j^s$.}

Second, we refine $\mathcal{G}_{\text{init}}$ using $C^{\text{coarse}}$ and $C^{\text{fine}}$ as complementary evidence. The MLLM processes $\mathcal{G}_{\text{init}}$ alongside these captions under structured instructions to verify and enrich the graph. Specifically, the model ensures mutual exclusion of contradictory actions, preserving the causal order of prerequisite and consequent events, and maintaining static object states across consecutive events unless a termination is explicitly captioned. Formally, this refinement process is defined as:
\begin{equation}
\mathcal{G}_{\text{final}} = f_{\text{MLLM}}(\mathcal{G}_{\text{init}}, C^{coarse}, C^{fine}),
\end{equation}
where $\mathcal{G}_{\text{final}}$ denotes the final verified and enriched Event-based Video Scene Graph. By filtering redundant information and mitigating textual hallucinations in the caption, the event graph provides a more reliable and fine-grained representation of event relations for video reasoning.

Fig.~\ref{fig:example} illustrates an example of EVSG. An EVSG is composed of \textit{event subgraphs} derived from multi-grained dense captions. 
EVSG converts caption cues into structured event triplets and aligns each event subgraph with start and end timestamps. Each event subgraph captures \textit{intra-event} entity-relation information, while timestamp-based edges encode temporal precedence between events for \textit{inter-event} reasoning. 
In this way, EVSG provides a concise and interpretable structure that improves key information selection for video reasoning, leading to more visually grounded, temporally consistent, and often stronger predictions than directly using captions.

\subsection{Event Graph-based Reinforcement Finetuning}
Our \ourmeth{} integrates the constructed EVSG into a GRPO-based reinforcement finetuning framework as a structured intermediate representation to guide and constrain reasoning. As illustrated in Fig.~\ref{fig:pipeline}, the MLLM takes as input the video $V$, its corresponding EVSG $\mathcal{G}_{\text{final}}$, and a query question $q$, forming a model output $o = [T, A] = \Phi(V, q, \mathcal{G}_{\text{final}})$. We optimise the model with a composite reward consisting of three components: 1) a format consistency reward $r_{\text{form}}(\cdot)$, 2) an accuracy reward $r_{\text{acc}}(\cdot)$, and 3) \textbf{a novel visual attention reward $r_{\text{attn}}(\cdot)$}.

\noindent \textbf{\textbullet~~Format Consistency Reward.}
To promote interpretable and structured reasoning, we introduce a template-based binary reward that verifies whether the output adheres to a predefined reasoning format. Specifically, the model must enclose intermediate reasoning within \texttt{<think>...</think>} tags and provide the final grounded response within \texttt{<answer>...</answer>} tags. The reward is defined as:
\begin{equation}
r_{\text{form}}(o) =
    \begin{cases}
    1, & \text{if $o$ satisfies the required format}, \\
    0, & \text{otherwise}.
    \end{cases}
\end{equation}
This constraint encourages stable reasoning traces and reduces degenerate or unstructured outputs during RL training.

\noindent \textbf{\textbullet~~Accuracy Reward.}
The accuracy reward evaluates both temporal localisation quality and semantic correctness. Given a model output $o$, we compute the temporal Intersection-over-Union (IoU) score with a sentence-level semantic similarity score:
\begin{equation}\label{eq:acc_reward}
r_{\text{acc}}(o) = \alpha \, r_{\text{sim}}(o) + (1 - \alpha) \, r_{\text{IoU}}(o),
\end{equation}
where $r_{\text{IoU}}$ measures the overlap between the predicted and ground-truth temporal intervals, and $r_{\text{sim}}$ computes semantic similarity between the generated answer and the reference text. The coefficient $\alpha \in [0,1]$ controls the trade-off between semantic correctness and temporal precision.

\noindent \textbf{\textbullet~~\textbf{Visual Attention Reward.}}
Although EVSG provides structured event relations, MLLMs may still over-rely on textual graph cues and under-utilise visual evidence, leading to visually ungrounded reasoning. To mitigate this drift, we introduce a visual attention reward that explicitly encourages response tokens to attend to video tokens.

Let $\mathbf{Attn}$ denote the averaged multi-head attention matrix across all layers, extracted from the output $o$ of MLLMs. We partition tokens into three groups: textual graph tokens $T_{\text{graph}}$, video tokens $T_{\text{vid}}$, and generated response tokens $T_{\text{res}}$. The attention reward is defined as the proportion of attention allocated from response tokens to video tokens relative to total tokens:
{\begin{equation}
\tilde r_{\text{attn}}(o) = 
\frac{\sum \mathbf{Attn}[T_{\text{res}}, T_{\text{vid}}]}
{\sum \mathbf{Attn}[T_{\text{res}}, T_{\text{vid}}] + \sum \mathbf{Attn}[T_{\text{res}}, T_{\text{graph}}]}.
\end{equation}}

A higher value indicates stronger reliance on visual evidence rather than abstract graph embeddings, promoting perceptually grounded reasoning. 
To prevent noisy optimisation signals, this reward is activated only when the generated output already satisfies a minimum reasoning quality threshold:
{\begin{equation}
r_{\text{attn}}(o) =
\begin{cases}
\tilde r_{\text{attn}}(o), & \text{if } r_{\text{sim}} \ge \tau_{\text{sim}} \text{ and } r_{\text{IoU}} \ge \tau_{\text{IoU}}, \\
0, & \text{otherwise},
\end{cases}
\end{equation}}
where $r_{\text{sim}}$ and $r_{\text{IoU}}$ are computed in Eq.~\ref{eq:acc_reward}, with $\tau_{\text{sim}}=0.4$ and $\tau_{\text{IoU}}=0.3$ based on ablation studies (in Fig.~\ref{fig:param_analysis}). This gated design ensures that attention optimisation only refines already reasonable predictions, avoiding destabilising low-quality samples. This design also mitigates potential bias introduced by EVSG construction, as the model is encouraged to ground its reasoning in video evidence rather than over-relying on graph-based textual cues.

\noindent \textbf{Overall Reward.}
The final reward is computed as a weighted combination:
\begin{equation}
r(o) = 
\lambda_{\text{form}} \, r_{\text{form}}(o)
+ \lambda_{\text{acc}} \, r_{\text{acc}}(o)
+ \lambda_{\text{attn}} \, r_{\text{attn}}(o),
\end{equation}
where $\lambda_{\text{acc}}, \lambda_{\text{form}}, \lambda_{\text{attn}}$ control the relative contributions of each component.  
This composite objective promotes semantic correctness, temporal grounding, structured reasoning, and visual evidence alignment, enabling the model to effectively leverage EVSG while reducing hallucinations during video reasoning.

\begin{table}[t]
\vspace{-10pt}
\caption{Comparison of SOTA methods on Rextime: ``ZS'', ``VLP'', ``SFT'', and ``RL'' denote Zero-shot, Vision-Language Pretrained, Supervised Fine-Tuning, and Reinforcement Learning. $^\dagger$ denotes results evaluated by us. ``--'' indicates not applicable. \shaded{The shaded} denote closed-source models.}
\centering
\renewcommand{\arraystretch}{1}
\setlength{\tabcolsep}{5pt}
\resizebox{1.0\linewidth}{!}{%
\begin{tabular}{l|cc|ccc|cc}
\toprule
\multirow{2}{*}{\textbf{Models}} & \multirow{2}{*}{\textbf{Param.}} & \multirow{2}{*}{\textbf{Type}} & \multicolumn{3}{c}{\textbf{Moment Localization}} & \multicolumn{2}{c}{\textbf{VQA}} \\
\cmidrule{4-6} \cmidrule{7-8}
 & & & mIoU & R@1,IoU=0.3 & R@1,IoU=0.5 & Acc. & Acc.@IoU$\geq$0.5 \\
\midrule
\rowcolor{black!6}
GPT-4o~\cite{OpenAIGPT4o} & -- & ZS & 36.28 & 45.33 & 34.00 & 73.67 & 28.67 \\
\rowcolor{black!6}
Claude3-Opus~\cite{anthropic2024claude} & -- & ZS & 23.61 & 30.67 & 17.67 & 68.67 & 13.67 \\
\rowcolor{black!6}
Gemini-1.5-Pro~\cite{team2023gemini} & -- & ZS & 28.43 & 35.67 & 25.00 & 68.00 & 18.33 \\
\rowcolor{black!6}
GPT-4V~\cite{OpenAIGPT4} & -- & ZS & 26.74 & 33.33 & 22.00 & 63.33 & 16.67 \\
\rowcolor{black!6}
\rowcolor{black!6}
Reka-Core~\cite{team2024reka} & -- & ZS & 27.95 & 36.33 & 24.00 & 59.67 & 17.00 \\
\midrule
UniVTG~\cite{lin2023univtg} & -- & VLP & 34.63 & 53.48 & 34.53 & -- & -- \\
CG-DETR~\cite{moon2023correlation} & -- & VLP & 26.53 & 39.71 & 22.73 & -- & -- \\
\midrule
VTimeLLM~\cite{huang2024vtimellm} & 7B & ZS & 20.14 & 28.84 & 17.41 & 36.16 & -- \\
TimeChat~\cite{ren2024timechat} & 7B & ZS & 11.65 & 14.42 & 7.61 & 40.04 & -- \\
LITA~\cite{huang2024lita} & 13B & ZS & 21.49 & 29.49 & 16.29 & 34.44& -- \\
TOGA~\cite{gupta2025toga} & 7B & ZS & 25.53 & 29.91 & 19.79 & -- & -- \\
VTimeLLM~\cite{huang2024vtimellm} & 7B & SFT & 29.92 & 43.69 & 26.13 & 57.58 & 17.13 \\
TimeChat~\cite{ren2024timechat} & 7B & SFT & 26.29 & 40.13 & 21.42 & 49.46 & 10.92 \\
TimeSearch~\cite{pan2025timesearch} & 7B & RL & 36.70 & 48.40 & 36.40 & 76.50 & 29.44 \\
VITAL ~\cite{zhang2025thinking} & 7B & RL & 40.90 & -- & -- & \textbf{79.10} & -- \\
\midrule
Qwen2.5-VL$^\dagger$~\cite{bai2025qwen2} & 7B & ZS & 13.60 & 16.05 & 9.24 & 56.60 & 6.35 \\
\ourmeth{}(w/o RL) & 7B & ZS & 25.34 & 33.92 & 20.25 & 66.82 & 15.21 \\
\ourmeth{} & 7B & RL & \textbf{41.46} & \textbf{57.54} & \textbf{40.36} & 71.30 & \textbf{30.75} \\
\bottomrule
\end{tabular}}
\vspace{-10pt}

\label{tab:rextime}
\end{table}

\section{Experiments}
\subsection{Experimental Setup}
\noindent \textbf{Datasets.}
We evaluate \ourmeth{} on two representative benchmarks that cover event-level causal reasoning and video temporal hallucination.
\textbf{(1) Rextime}~\cite{chen2024rextime} is a grounded VQA benchmark centred on event causal relations. It is designed to evaluate complex temporal reasoning, requiring models to answer causal questions while providing temporally consistent localisation. The dataset contains 9,695/921/2,143 samples for the train/val/test splits, respectively.
\textbf{(2) VidHalluc}~\cite{li2025vidhalluc} is an evaluation-only benchmark for video temporal hallucination. It comprises 5,002 videos and 9,295 questions, spanning three critical dimensions of video hallucination: \emph{Action Hallucination (ACH)}, \emph{Temporal Sequence Hallucination (TSH)}, and \emph{Scene Transition Hallucination (STH)}.

\noindent \textbf{Implementation Details.}
We use Qwen2.5-VL-7B~\cite{bai2025qwen2} as both the reasoning backbone and the EVSG constructor for all main experiments. The event-number constraints are set to $\{5,10,15\}$. For GRPO finetuning, we use reward weights $\lambda_{\text{acc}}=0.7$, $\lambda_{\text{form}}=0.3$, $\lambda_{\text{attn}}=0.6$, and $\alpha=0.3$, following the ablations in Fig.~\ref{fig:param_analysis}. All experiments are implemented in PyTorch and conducted on 8 NVIDIA A100 GPUs, with 8 rollouts, batch size 16, and one training epoch. More implementation details and prompts are provided in the appendix.

\noindent \textbf{Evaluation Metrics.}
For RexTime, we report answer Accuracy (Acc), grounded accuracy Acc@IoU$\geq$0.5 (Acc@0.5), R@1 at IoU thresholds 0.3/0.5, and mIoU. Acc@0.5 jointly measures answer correctness and temporal grounding by requiring both a correct answer and a predicted temporal span with at least 0.5 IoU with the ground truth. For \textbf{VidHalluc}, we follow the official benchmark protocol. Specifically, we report \emph{Accuracy} for the Action Hallucination (ACH) and Temporal Sequence Hallucination (TSH) subsets, and the official \emph{Score} for the Scene Transition Hallucination (STH) subset.

\begin{table}[t]
\vspace{-10pt}
\caption{Comparisons of methods on the three test sets on the VidHalluc benchmark, including Action Hallucination (ACH), Temporal Sequence Hallucination (TSH) and Scene Transition Hallucination (STH) test. $^\dagger$ denotes results evaluated by us. *denotes finetuned on Rextime. \shaded{The shaded} denote closed-source models.}
\centering
\small
\renewcommand{\arraystretch}{1.1}
\setlength{\tabcolsep}{5pt}
\resizebox{0.9\linewidth}{!}{%
\begin{tabular}{l|c|cccc}
\toprule
\multirow{2}{*}{\textbf{Models}} & \multirow{2}{*}{\textbf{Params}} & \multicolumn{2}{c}{\textbf{Accuracy on ACH$\uparrow$}} & \multirow{2}{*}{\shortstack{\textbf{Accuracy} \\ \textbf{on TSH$\uparrow$}}} & \multirow{2}{*}{\shortstack{\textbf{Score} \\ \textbf{on STH$\uparrow$}}} \\
\cmidrule{3-4}
 &  & Binary QA$\uparrow$ & MCQ$\uparrow$ &  &  \\
\midrule
\rowcolor{black!6}
Gemini-1.5-Pro~\cite{team2023gemini} & $\sim$200B & 75.27 & 79.25 & 83.83 & 63.96 \\
\rowcolor{black!6}
GPT-4o~\cite{OpenAIGPT4o} & $\sim$200B & \textbf{81.15} & \textbf{90.95} & \textbf{82.00} & \textbf{71.58} \\
\midrule
Video-ChatGPT~\cite{maaz2024video} & 7B & 9.50 & 24.58 & 30.17 & 7.70 \\
Video-LLAVA~\cite{lin2024video} & 7B & 26.84 & 64.45 & 27.17 & 29.60 \\
ShareGPT4Video~\cite{chen2024sharegpt4video} & 8B & 29.96 & 44.78 & 49.50 & 17.83 \\
Chat-UniVi~\cite{jin2024chat} & 13B & 23.77 & 54.79 & 35.50 & 29.87 \\
LLaVA-NeXT-Video~\cite{zhang2024llava} & 34B & 26.60 & 77.57 & 21.33 & 44.40 \\
PLLaVA~\cite{xu2024pllava} & 13B & 35.30 & 76.96 & 16.50 & 32.44 \\
VideoLLaMA2~\cite{cheng2024videollama} & 7B & 50.04 & 83.84 & 26.17 & \textbf{65.12} \\
VILA1.5~\cite{lin2024vila} & 13B & 58.46 & 81.88 & 63.33 & 35.03 \\
\midrule
Qwen2.5-VL$^\dagger$~\cite{bai2025qwen2} & 7B & 50.46 & 81.99 & 67.50 & 46.77 \\
\ourmeth{}(w/o RL) & 7B & 50.92 & 83.23 & 75.33 & 54.58 \\
\ourmeth{}* & 7B & \textbf{66.04} & \textbf{84.57} & \textbf{76.33} & 57.81 \\
\bottomrule
\end{tabular}}

\label{tab:vidhalluc}
\vspace{-15pt}
\end{table}

\subsection{Comparison with SoTAs}
\noindent \textbf{Results on RexTime.}
Tab.~\ref{tab:rextime} compares \ourmeth{} with state-of-the-art methods on RexTime. In the training-free setting, incorporating EVSG into the baseline model already brings notable gains in reasoning-grounded VQA, improving mIoU by $11.74\%$, Accuracy by $10.22\%$, and Accuracy@IoU$\geq$0.5 by $8.86\%$. This suggests that EVSG provides useful representation for temporally grounded reasoning beyond dense caption inputs.
With event graph-based RL post-training, \ourmeth{} further improves localisation and achieves competitive VQA accuracy compared with tool-augmented methods such as VITAL~\cite{zhang2025thinking} and TimeSearch~\cite{pan2025timesearch}. While these methods segment long videos into short clips, \ourmeth{} reasons over the full video with EVSG context, leading to stronger Accuracy@IoU$\geq$0.5 and better temporal alignment. Notably, \ourmeth{} also surpasses GPT-4o by $2.08\%$ on Accuracy@IoU$\geq$0.5, indicating reinforcement finetuning enables the model to better leverage EVSG for event-level reasoning, leading to more consistent temporal localisation.

\noindent \textbf{Results on VidHalluc.}
Tab.~\ref{tab:vidhalluc} evaluates video temporal hallucination on VidHalluc. In the training-free setting, integrating EVSG consistently improves Qwen2.5-VL across all dimensions, with gains of $7.83\%$ on Temporal Sequence Hallucination (TSH) and $7.81\%$ on Scene Transition Hallucination (STH). These results suggest that explicit event structure helps reduce hallucination in temporal errors. Since VidHalluc is an evaluation-only benchmark without a training split, \ourmeth{} is directly evaluated without VidHalluc-specific finetuning. After GRPO finetuning on RexTime, \ourmeth{} further improves its ability to use EVSG for grounded reasoning and achieves substantial gains across all dimensions. Although closed-source models such as GPT-4o and Gemini-1.5-Pro remain stronger overall, \ourmeth{} narrows the gap with a compact 7B-scale backbone, showing the effectiveness of structured event modelling and reinforcement finetuning.
\begin{wrapfigure}{r}{0.45\textwidth}
\captionsetup{type=table}
\vspace{35pt}
\caption{Ablation studies of Qwen2.5-VL-7B on Rextime validation set. DC: dense captions.}
\vspace{-5pt}
\centering
\renewcommand{\arraystretch}{1.1}
\resizebox{\linewidth}{!}{%
\begin{tabular}{lccc}
\toprule
\textbf{Methods} & \textbf{mIoU} & \textbf{Acc} & \textbf{Acc@0.5} \\
\midrule
\rowcolor[gray]{0.9}
\multicolumn{4}{l}{\textit{\textbf{Training-free}}} \\
Qwen2.5-VL-7B & 13.77 & 55.65 & 6.20 \\
+ DC & 24.25 & 62.87 & 15.42 \\
+ EVSG & \textbf{27.98} & \textbf{64.17} & \textbf{18.02} \\
\midrule
\rowcolor[gray]{0.9}
\multicolumn{4}{l}{\textit{\textbf{GRPO Finetuned}}} \\
Qwen2.5-VL-7B & 41.04 & 57.55 & 27.03 \\
+ DC & 41.20 & 69.60 & 29.10 \\
+ DC + $r_{attn}$ & 41.42 & 71.01 & 32.72 \\
+ EVSG & 42.57 & 72.20 & 32.03 \\
+ EVSG + $r_{attn}$ & \textbf{43.56} & \textbf{73.72} & \textbf{33.49} \\
\bottomrule
\end{tabular}%
}

\label{tab:ablation_qwenvl}
\vspace{-40pt}
\end{wrapfigure}

\subsection{Ablation Studies}
\noindent \textbf{Effect of using EVSG.}
We conducted ablation studies with the Qwen2.5-VL-7B model on the validation set of the Rextime benchmark to systematically evaluate the contribution of each component in our approach. 
In Tab.~\ref{tab:ablation_qwenvl}, we first conduct training-free results in rows 1-3. The results show that adding dense captions (DC) already improves over the base model by providing extra visual cues in text, but the cues remain buried in long, unstructured text, where causal steps are easily distracted by redundant descriptions and hard to consistently retrieve for temporal grounding. EVSG further addresses this bottleneck by event-level entities and relations and explicitly connecting them with temporal links, making both intra-event interactions and inter-event dependencies directly accessible during reasoning. Consequently, EVSG yields more coherent semantics and more accurate localisation, improving Acc@0.5 from 15.42\% to 18.0\% compared to DC (from 6.20\% to 18.02\%).

\begin{figure}[t]
\centering
\includegraphics[width=\columnwidth]{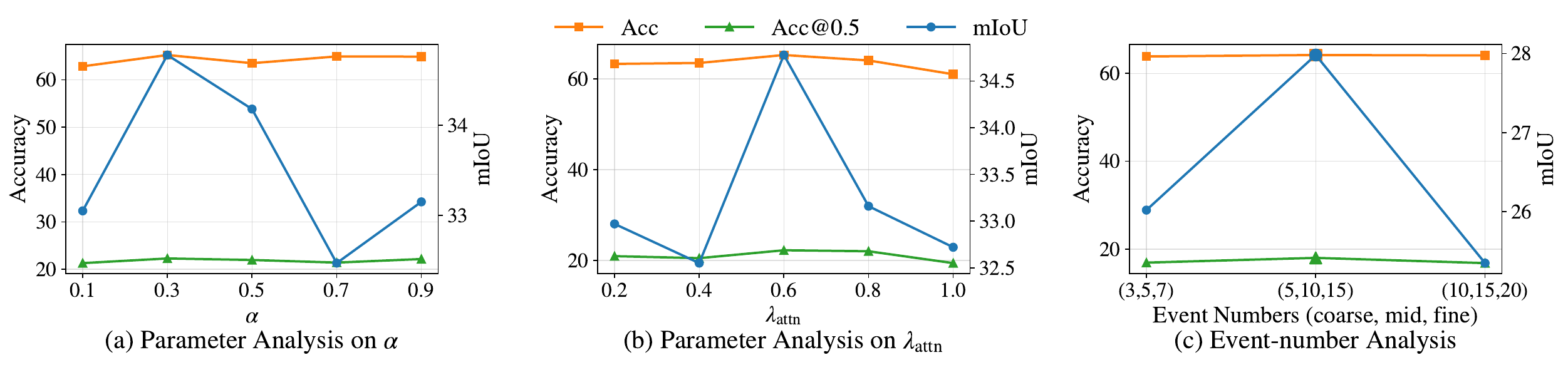}
\vspace{-15pt}
\caption{Parameter sensitivity on the Rextime validation set.}
\vspace{-10pt}
\label{fig:param_analysis}
\end{figure}

\begin{figure}[t]
   \centering
   %
   \begin{subfigure}[t]{0.23\columnwidth}
      \centering
      \includegraphics[width=\linewidth]{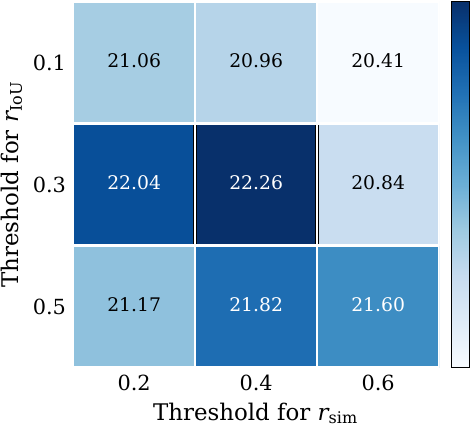}
      \caption{Threshold sensitivity.}
      \label{fig:threshold_analysis}
   \end{subfigure}
   \hfill
   \begin{subfigure}[t]{0.38\columnwidth}
      \centering
      \includegraphics[width=\linewidth]{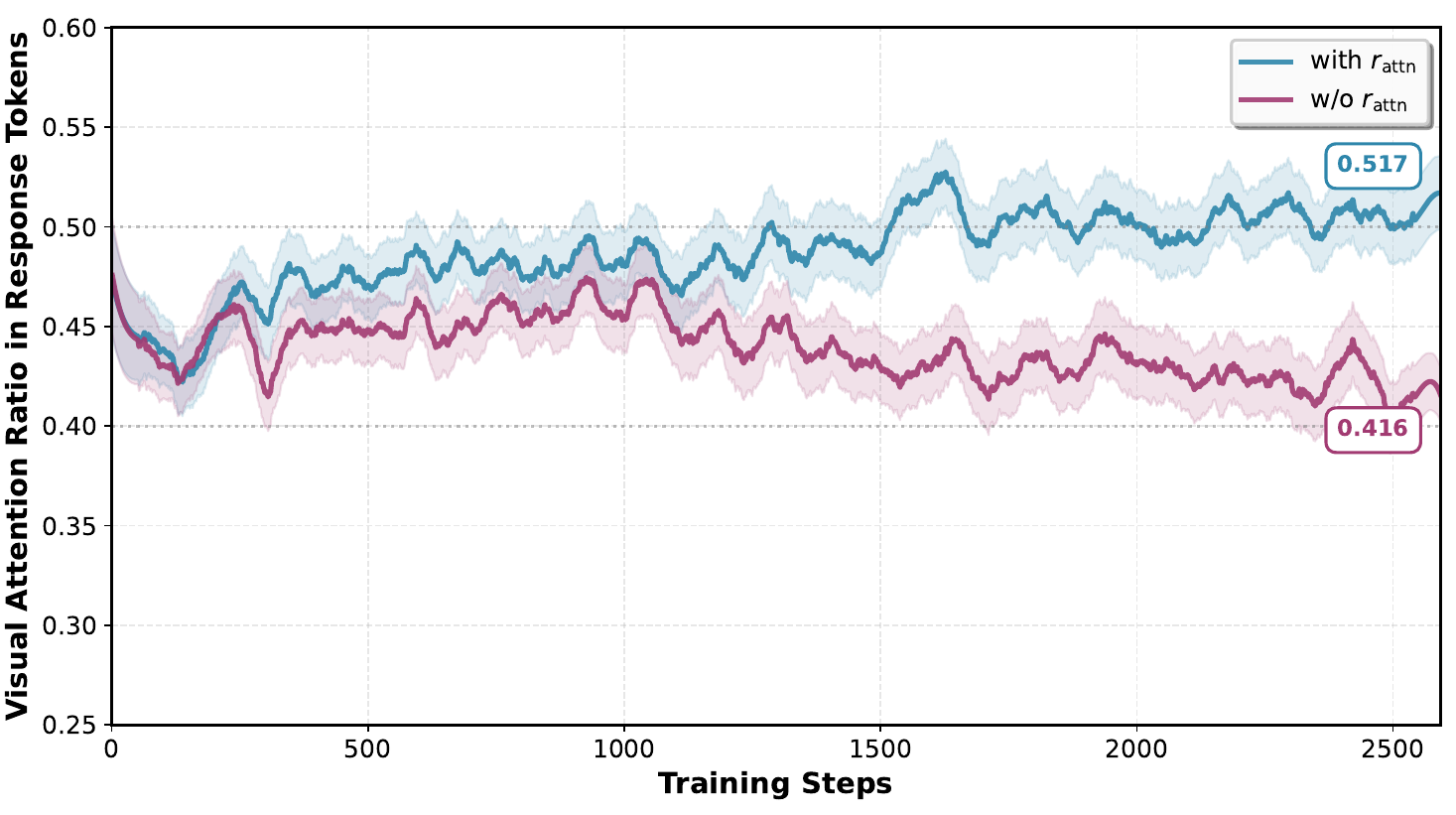}
      \caption{Impact of $r_{\text{attn}}$ on visual attention.}
      \label{fig:attn_visual_impact}
   \end{subfigure}
   \hfill
   \begin{subfigure}[t]{0.32\columnwidth}
      \centering
      \centering
      \includegraphics[width=\linewidth]{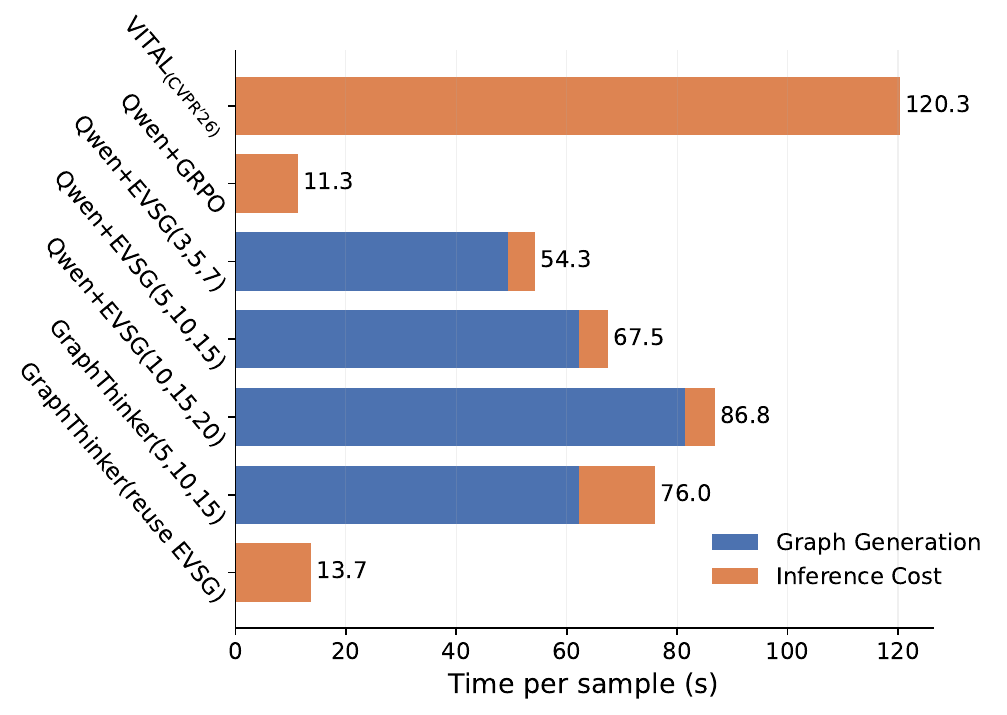}
      \caption{Inference cost (sec/query).}
      \label{fig:efficiency_analysis}
   \end{subfigure}
   \vspace{-5pt}
   \caption{(a) Threshold sensitivity heatmap for $r_{\text{sim}}$ and $r_{\text{IoU}}$, reporting Acc@0.5 on the RexTime validation set with Qwen2.5-VL-3B. (b) Effect of the attention reward $r_{\text{attn}}$ on visual attention. (c) Inference cost analysis.}
   \label{fig:threshold_attn_comparison}
   \vspace{-15pt}
\end{figure}

\noindent \textbf{Effect of RL Finetuning.}
Moreover, our EVSG demonstrates superior event relation modelling compared to dense captions under RL finetuning. Applying GRPO alone improves performance over the training-free settings, while combining it with EVSG and the visual attention reward $r_{attn}$ yields further gains across all metrics, increasing mIoU, Acc, and Acc@0.5 by $2.52\%$, $16.17\%$, and $6.46\%$, respectively. 
These results clearly show that the baseline (Qwen2.5-VL-7B) benefits substantially from EVSG, even without additional finetuning. Moreover, RL finetuning with EVSG and $r_{\text{attn}}$ consistently improves performance on both GRPO without textual supplements and GRPO using only dense captions as the intermediate representation. These comparisons prove the EVSG's effectiveness in both training-free and fine-tuned settings.

\vspace{-5pt}
\subsection{Further Analysis}

\noindent \textbf{Reward Weights.}
Fig.~\ref{fig:param_analysis}(a)/(b) analyses the effects of $\alpha$, $\lambda_{\text{attn}}$
in the reward function. The results show that performance peaks at $\alpha=0.3$ (balancing $r_{\text{sim}}$ and $r_{\text{IoU}}$) and $\lambda_{\text{attn}}=0.6$, which moderately regularises visual attention. This suggests that effective post-training requires a balanced reward design that jointly considers answer correctness, temporal grounding, and visual-token reliance.

\noindent \textbf{Effect of Event Numbers.}
Fig.~\ref{fig:param_analysis}(c) studies the impact of event numbers in multi-grained caption generation. Since most videos in our experiments are around 3 mins long, we compare different granularities for constructing EVSGs. Too few events omit important details, while too many events make captions redundant or noisy. The setting $(5,10,15)$ provides a better trade-off between detail and compactness, leading to more coherent and informative event graphs.

\noindent \textbf{Analysis of the Visual-attention Reward.}
Fig.~\ref{fig:threshold_attn_comparison} analyses the reward design from two aspects. Fig.~\ref{fig:threshold_analysis} shows that the selected thresholds (chosen by the best RexTime validation performance) achieve the best result and lie in a relatively stable region, suggesting a good trade-off between reward effectiveness and noise. Fig.~\ref{fig:attn_visual_impact} further shows that adding $r_{\text{attn}}$ consistently increases the visual attention ratio during training and leads to a higher final ratio (0.517 vs.\ 0.416), indicating better visually grounded reasoning.

\noindent \textbf{Inference Efficiency.}
Fig.~\ref{fig:efficiency_analysis} reports the average inference time per query under different settings. The training-free Qwen2.5-VL-7B + EVSG variants first construct EVSGs from multi-grained captions and then perform graph-conditioned inference. Based on the accuracy-efficiency trade-off in Fig.~\ref{fig:param_analysis}(c), we use the $(5,10,15)$ event setting for \ourmeth{}. Although EVSG construction introduces additional first-query cost, \ourmeth{} achieves 76.0s per query, compared with 120.3s for the tool-based SoTA VITAL~\cite{zhang2025thinking}, which segments long videos during inference. Moreover, EVSGs can be reused for multiple queries on the same video, making later queries substantially cheaper and close to the vanilla GRPO-trained Qwen model (as shown at the bottom of Fig.~\ref{fig:efficiency_analysis}).

\begin{figure}[t]
  \centering
  \includegraphics[width=1.\linewidth]{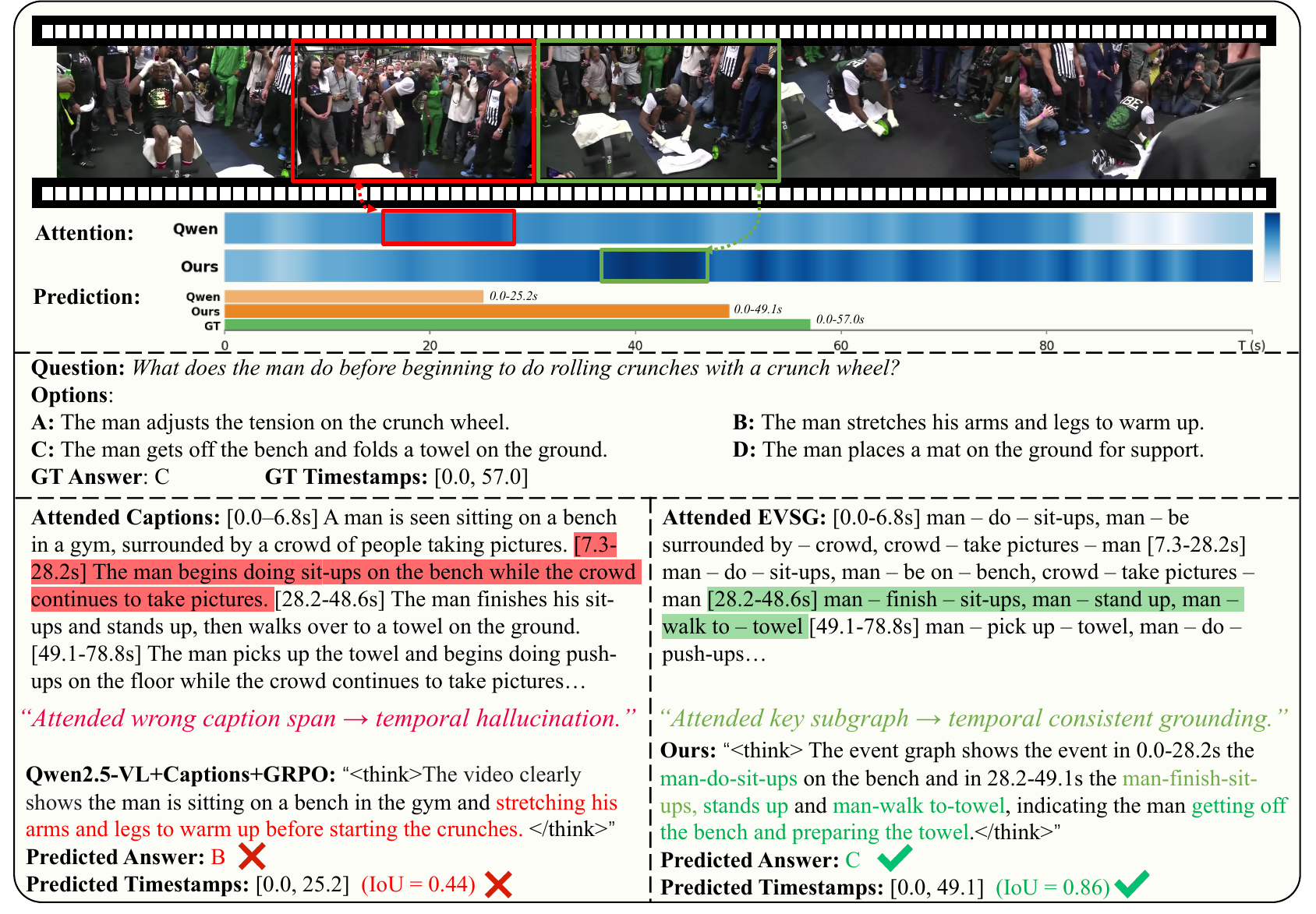}
  \vspace{-15pt}
  \caption{Illustration of \ourmeth{} reducing temporal hallucinations in reasoning. Qwen2.5-VL with dense captions after GRPO still produces hallucinated predictions, as redundant captions can obscure key evidence. In contrast, \ourmeth{} uses EVSG to organise key entities, relations, and their temporal order, guiding the model to focus on the relevant visual cues and produce more visually grounded and temporally consistent predictions. The visual attention reward further encourages attention to related video tokens during reasoning.
  }
  \label{fig:visualization}
  \vspace{-15pt}
\end{figure}

\noindent \textbf{Qualitative Examples.}
Fig.~\ref{fig:visualization} illustrates an example demonstrating that \ourmeth{} mitigates temporal hallucinations in video reasoning. With caption-based GRPO, Qwen2.5-VL misinterprets the scene as a warm-up activity (\textcolor{red}{Option B}) and exhibits temporal sequence hallucination by incorrectly inferring event order. In contrast, \ourmeth{} leverages EVSG to represent object and event relations more explicitly, capturing that the man performs sit-ups, gets off the bench, folds a towel, and then begins rolling crunches. As a result, it predicts the correct answer (\textcolor{green}{Option C}) and achieves more accurate temporal localisation ([0.0--49.1s], IoU = 0.86) in reasoning. This example illustrates that explicit event structure can reduce sequential hallucination, while the visual attention reward encourages the model to place greater attention on relevant video tokens during reasoning.


\section{Conclusion}

This work introduces \ourmeth{}, a reinforcement finetuning framework that improves temporally grounded video reasoning for MLLMs. By constructing an Event-based Video Scene Graph (EVSG) as a structured intermediate representation, \ourmeth{} explicitly models intra-event entity relations and inter-event temporal relations without manual annotations. We further integrate EVSG into the GRPO framework with a novel visual attention reward to strengthen visual grounding during video reasoning. Together, structured event modelling and attention reward reduce hallucinations and enhance temporal consistency. Extensive experiments demonstrate that \ourmeth{} achieves state-of-the-art performance on RexTime and VidHalluc.

\bibliography{main}
\bibliographystyle{plain}

\newpage
\appendix

\section{Additional Implementation Details}
\label{app_sec1}
This section provides further implementation details, including the MLLM configurations, the prompt templates used for EVSG construction, and the prompt templates used for evaluation on the ReXTime~\cite{chen2024rextime} benchmark.

\subsection{MLLM settings}
\label{app:settings}
In the experiments, we used Qwen2.5-VL-7B~\cite{bai2025qwen2} as the baseline model for GRPO finetuning. To ensure that models receive sufficient video information for reasoning, the video input sampling rate was set to 1 FPS for both the ReXTime~\cite{chen2024rextime} and VidHalluc~\cite{li2025vidhalluc} benchmarks. The maximum prompt length was set to 4096 tokens, with a response length of 2048 tokens, 8 rollouts, a batch size of 16, and 1 training epoch. In the parameter analysis, the same settings were applied to the Qwen2.5-VL-3B model. When using Qwen2.5-VL-3B for evaluation, the EVSG graphs were also generated by Qwen2.5-VL-3B. The Sentence-Transformer model used to compute sentence similarity in Eq.~(6) of the main paper is all-MiniLM-L6-v2~\cite{wang2020minilm}.

\subsection{Prompt Templates for EVSG Construction}
\label{app:graph_prompt}
We first generate multi-grained dense captions, and subsequently initialise and refine EVSGs based on them.

\begin{figure}[!h]
   \centering 
   \includegraphics[width=0.94\textwidth]{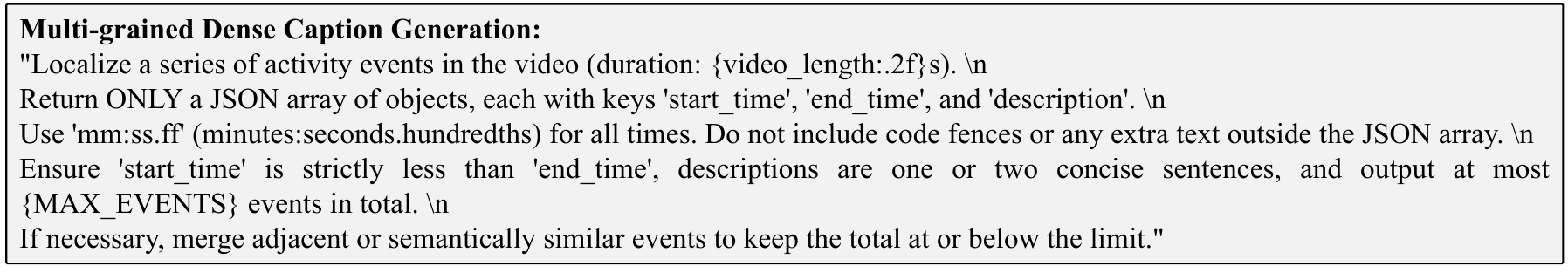}
   \caption{The prompt template for multi-grained dense caption generation.}
   \label{fig:prompt_cap}
\end{figure}

\noindent\textbf{Caption generation.} Fig.~\ref{fig:prompt_cap} illustrates the prompt template for generating multi-grained dense captions. The predefined event number limits replace the placeholder \texttt{MAX\_EVENTS} to constrain the maximum number of events during multi-grained dense caption generation.

\begin{figure}[!h]
   \centering 
   \includegraphics[width=0.94\textwidth]{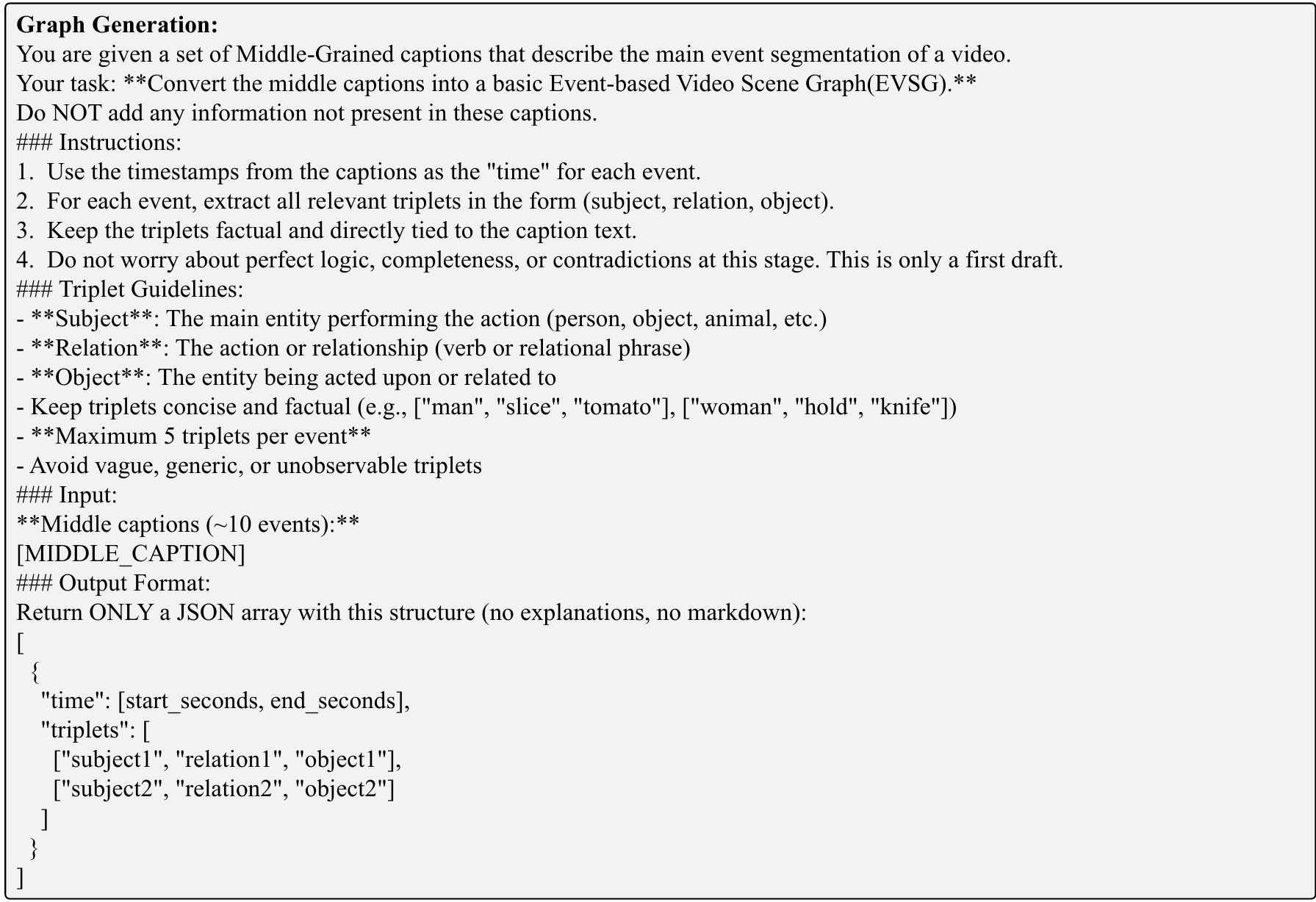}
   \caption{The prompt template for graph generation with middle-grained dense captioning.}
   \label{fig:prompt_graph1}
\end{figure}

\noindent\textbf{Graph initialisation.} The prompt template for graph generation is shown in Fig.~\ref{fig:prompt_graph1}. Following these instructions, the used MLLM generates a graph draft in JSON format based on the middle-grained dense captions. The placeholder \texttt{MIDDLE\_CAPTION} is replaced by the middle-grained captions produced in the preceding step.

\begin{figure}[!h]
   \centering 
   \includegraphics[width=0.94\textwidth]{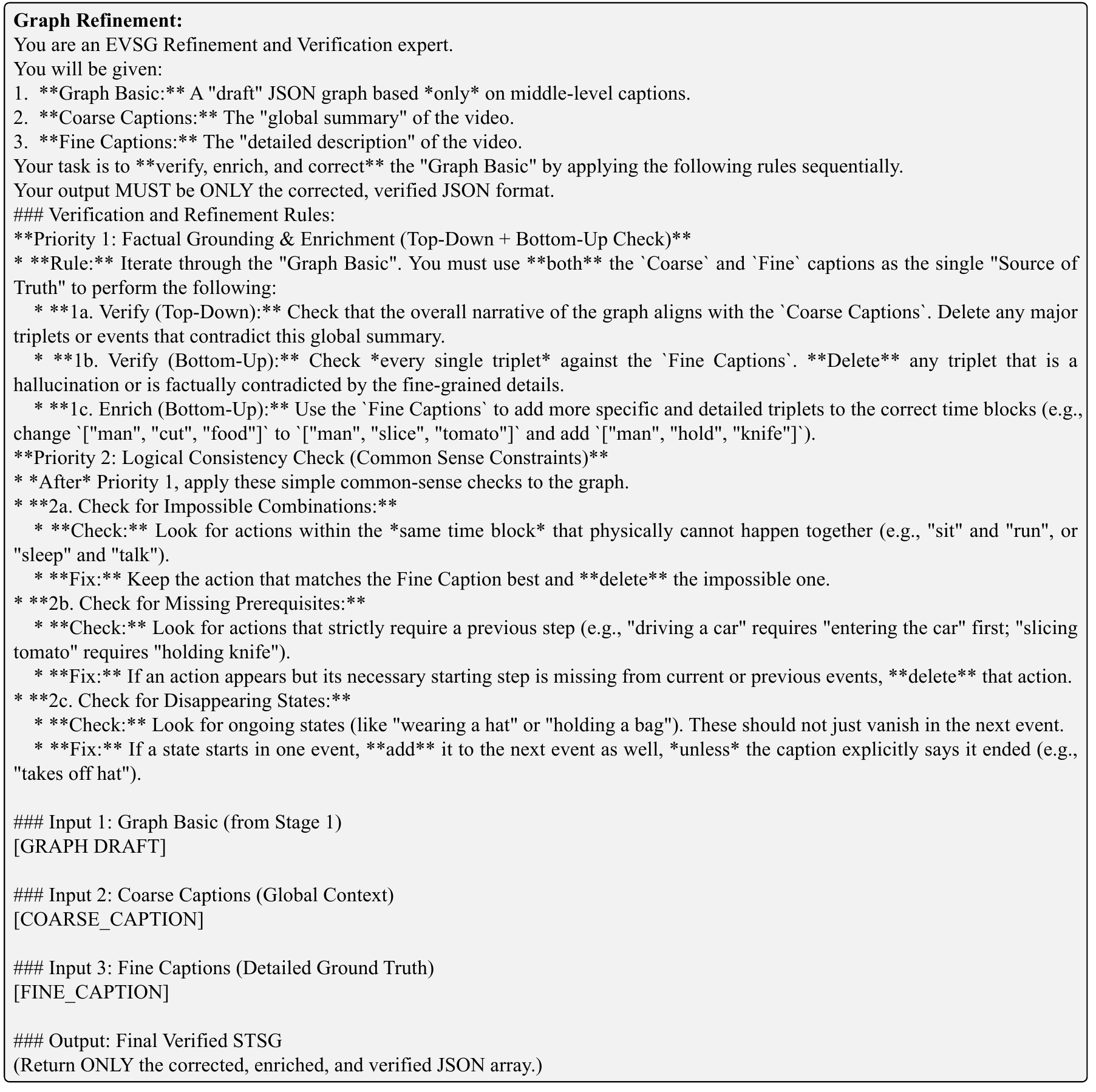}
   \caption{The prompt template for graph refinement with coarse- and fine-grained dense captioning.}
   \label{fig:prompt_graph2}
\end{figure}

\noindent\textbf{Graph refinement.} Finally, Fig.~\ref{fig:prompt_graph2} presents the prompt template for graph refinement. The placeholders \texttt{GRAPH\_DRAFT}, \texttt{COARSE\_CAPTION}, and \texttt{FINE\_CAPTION} are replaced by the graph and captions generated in the previous steps.

\subsection{Prompt Templates for Evaluation}
\label{app:eval_prompt}
\begin{figure}[!h]
   \centering 
   \includegraphics[width=0.94\textwidth]{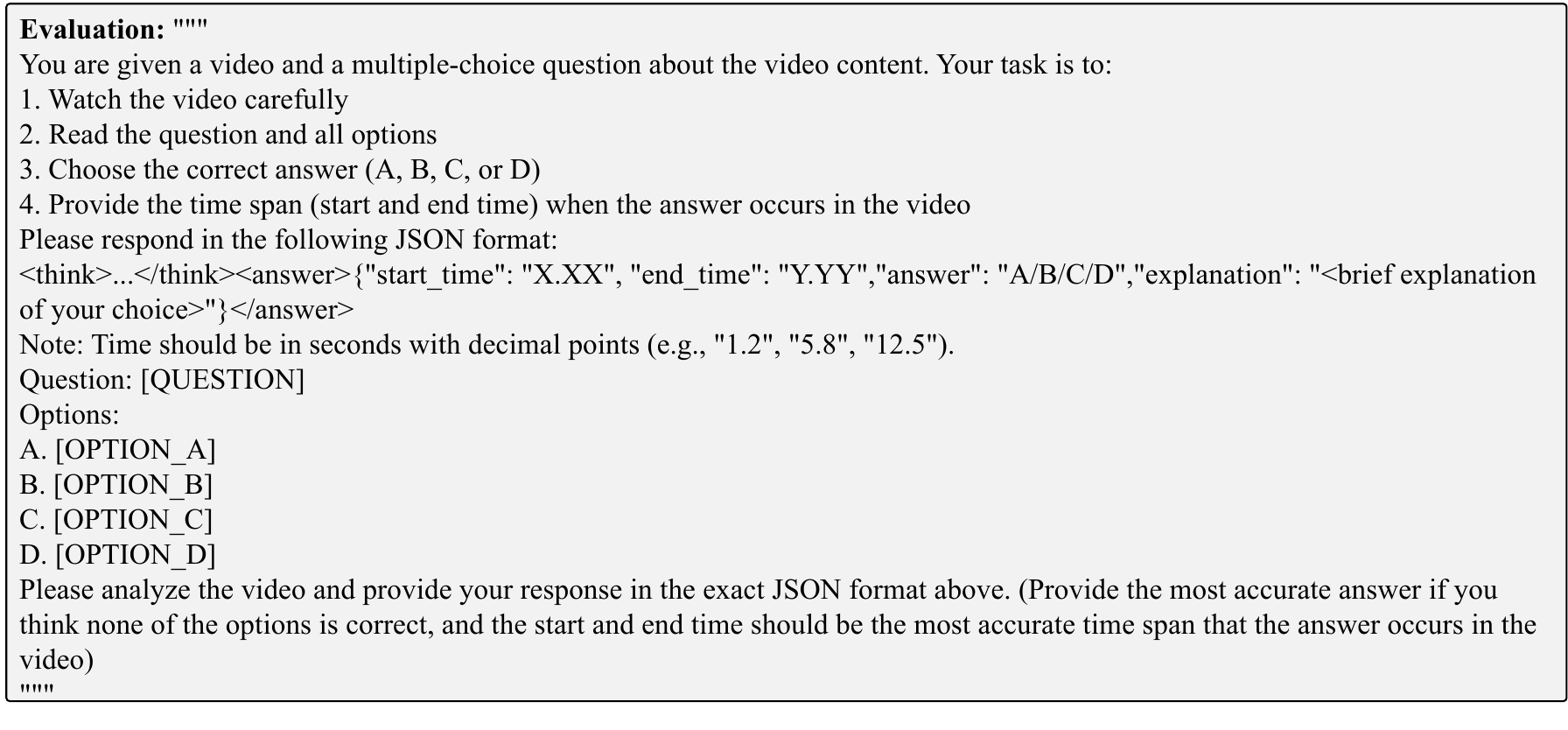}
   \vspace{-10pt}
   \caption{The prompt template for evaluation on the ReXTime~\cite{chen2024rextime} benchmark.}
   \label{fig:prompt_eval}
\vspace{-10pt}
\end{figure}

Fig.~\ref{fig:prompt_eval} shows the prompt template used for \ourmeth{} in the evaluation on the ReXTime~\cite{chen2024rextime} benchmark. The response format is identical to that used during GRPO fine-tuning, ensuring that the model consistently follows the instructions and produces outputs in the correct format at evaluation time. For VidHalluc~\cite{li2025vidhalluc} evaluation, we follow the prompt template provided by the benchmark.

\section{Additional Ablation Studies}
\label{app_sec2}
\subsection{Effect of Caption Granularity}
\label{app:graph_granularity}
Tab.~\ref{tab:tb3} presents a further analysis of the impact of using different-grained captions for EVSG construction. If we only use single-granularity captions, the generated middle-grained EVSG achieves the best Acc@0.5 score of 17.48, suggesting that a moderate level of event granularity provides the most effective balance between semantic coverage and temporal precision. The generated coarse-grained graph, whilst more compact, tends to overlook fine-grained temporal details, leading to weaker localisation performance. The generated fine-grained graph introduces richer event descriptions and yields a marginally higher Acc of 63.74, but its lower mIoU and Acc@0.5 indicate that excessively dense event decomposition can introduce redundancy and undermine temporal consistency. Notably, the generated EVSGs that combine captions across all granularities consistently achieve the best performance across all metrics, reaching an mIoU of 27.98, an Acc of 64.17, and an Acc@0.5 of 18.02. These results demonstrate that integrating coarse-, middle-, and fine-grained evidence enables the EVSGs to capture complementary temporal and semantic information, thereby producing a more comprehensive and discriminative event representation for video reasoning and grounding.
\begin{table}[t]
\caption{Comparisons of using different-grained captions for EVSG generation. }
\centering
\begin{tabular}{c|ccc}
\hline
EVSG (Event Number) & mIoU & Acc & Acc@0.5 \\ \hline
Coarse-grained (5) & 25.76 & 62.54 & 16.94 \\
Middle-grained (10) & 26.60 & 62.76 & 17.48 \\
Fine-grained (15) & 24.98 & 63.74 & 16.83 \\
Multi-grained (5,10,15) & 27.98 & 64.17 & 18.02 \\ \hline
\end{tabular}
\label{tab:tb3}
\end{table}

\subsection{Effect of Graph Constructor Quality}
\label{app:constructor_quality}

To analyse whether the quality of EVSG construction affects downstream reasoning, we compare different graph constructors while keeping the reasoning backbone fixed as Qwen2.5-VL-7B. Specifically, we use either Qwen2.5-VL-7B or Gemini-2.5-Flash to generate dense captions and EVSGs, and then feed the resulting context to the same Qwen2.5-VL-7B baseline.

As shown in Tab.~\ref{tab:constructor_analysis}, EVSG consistently outperforms dense captions under both constructors. With Qwen2.5-VL-7B as the constructor, EVSG improves Acc@0.5 from 15.42 to 18.02. With Gemini-2.5-Flash as the constructor, EVSG further improves Acc@0.5 from 20.41 to 21.93. These results suggest that the benefit of EVSG does not merely come from adding extra textual descriptions, but from organising video evidence into an event-structured representation.

We also observe that using Gemini-2.5-Flash as the constructor improves both dense captions and EVSGs, indicating that higher-quality captioning and graph construction can further benefit GraphThinker. This suggests that the proposed framework is not inherently tied to Qwen2.5-VL-7B as the graph constructor and can potentially benefit from stronger MLLMs.

Nevertheless, for the main experiments, we use Qwen2.5-VL-7B as both the reasoning baseline and the EVSG constructor. This choice avoids introducing an external stronger model into the main comparison and provides a controlled open-source setting for evaluating the effect of EVSG itself. The Gemini-based results are therefore used only as an analysis of constructor quality rather than as the main reported setting.
\begin{table}[t]
\centering
\caption{Effect of different graph constructors on RexTime validation set. The baseline is fixed as Qwen2.5-VL-7B. ``DC'' denotes dense captions.}
\label{tab:constructor_analysis}
\resizebox{0.60\linewidth}{!}{
\begin{tabular}{llccc}
\toprule
\textbf{Constructor} & \textbf{Context} & \textbf{mIoU} & \textbf{Acc} & \textbf{Acc@0.5} \\
\midrule
Qwen2.5-VL-7B    & DC   & 24.25 & 62.87 & 15.42 \\
Qwen2.5-VL-7B    & EVSG & 27.98 & 64.17 & 18.02 \\
Gemini-2.5-Flash & DC   & 26.78 & 71.55 & 20.41 \\
Gemini-2.5-Flash & EVSG & \textbf{28.92} & \textbf{72.64} & \textbf{21.93} \\
\bottomrule
\end{tabular}
}
\end{table}

\subsection{Faithfulness Analysis of EVSG}
\label{app:evsg_faithfulness}

Since EVSG is generated by an MLLM, we conduct a small-scale manual audit to examine whether the generated graph faithfully reflects the video content. We randomly sample 50 videos from the RexTime validation set and inspect the generated EVSG together with the original video. We evaluate two key aspects: event boundary correctness and triplet faithfulness. Event boundary correctness checks whether the generated event nodes are aligned with the corresponding temporal segments, while triplet faithfulness checks whether the key subject-relation-object triplets are visually supported by the corresponding video segment. 

Each aspect is rated as Correct, Partial, or Incorrect. Correct means that the aspect is largely consistent with the video. Partial means that the main information is correct but contains minor ambiguity, omission, or weakly supported details. Incorrect means that the aspect contains major errors that may mislead reasoning. For event boundaries, errors include over-segmentation, under-segmentation, or assigning an event to the wrong temporal region. For triplets, errors include unsupported or contradictory subject-relation-object relations in the corresponding video segment.

\begin{table}[t]
\centering
\caption{Sample-level manual faithfulness audit of EVSG on 50 randomly sampled RexTime validation videos. Each aspect is rated as Correct, Partial, or Incorrect for the whole generated EVSG of each sample.}
\label{tab:evsg_faithfulness}
\resizebox{0.85\linewidth}{!}{
\begin{tabular}{lccc}
\toprule
\textbf{Aspect} & \textbf{Correct (\%)} & \textbf{Partial (\%)} & \textbf{Incorrect (\%)} \\
\midrule
Event boundary correctness & 76 & 18 & 6 \\
Triplet faithfulness & 72 & 16 & 12 \\
\bottomrule
\end{tabular}
}
\end{table}

As shown in Tab.~\ref{tab:evsg_faithfulness}, most EVSGs provide reasonable event segmentation and visually supported relation triplets, suggesting that EVSG can serve as useful structured context for temporally grounded reasoning. However, since EVSG is generated by an MLLM, it may still contain boundary errors or unsupported triplets. Therefore, we do not treat EVSG as ground-truth supervision, but use it as a structured intermediate representation while retaining the original video input for grounding.

\section{More Qualitative Analysis}
\label{app:qualitative}
We show more qualitative analysis in this section.

\begin{figure}[!h]
  \centering
  \includegraphics[width=1.0\linewidth]{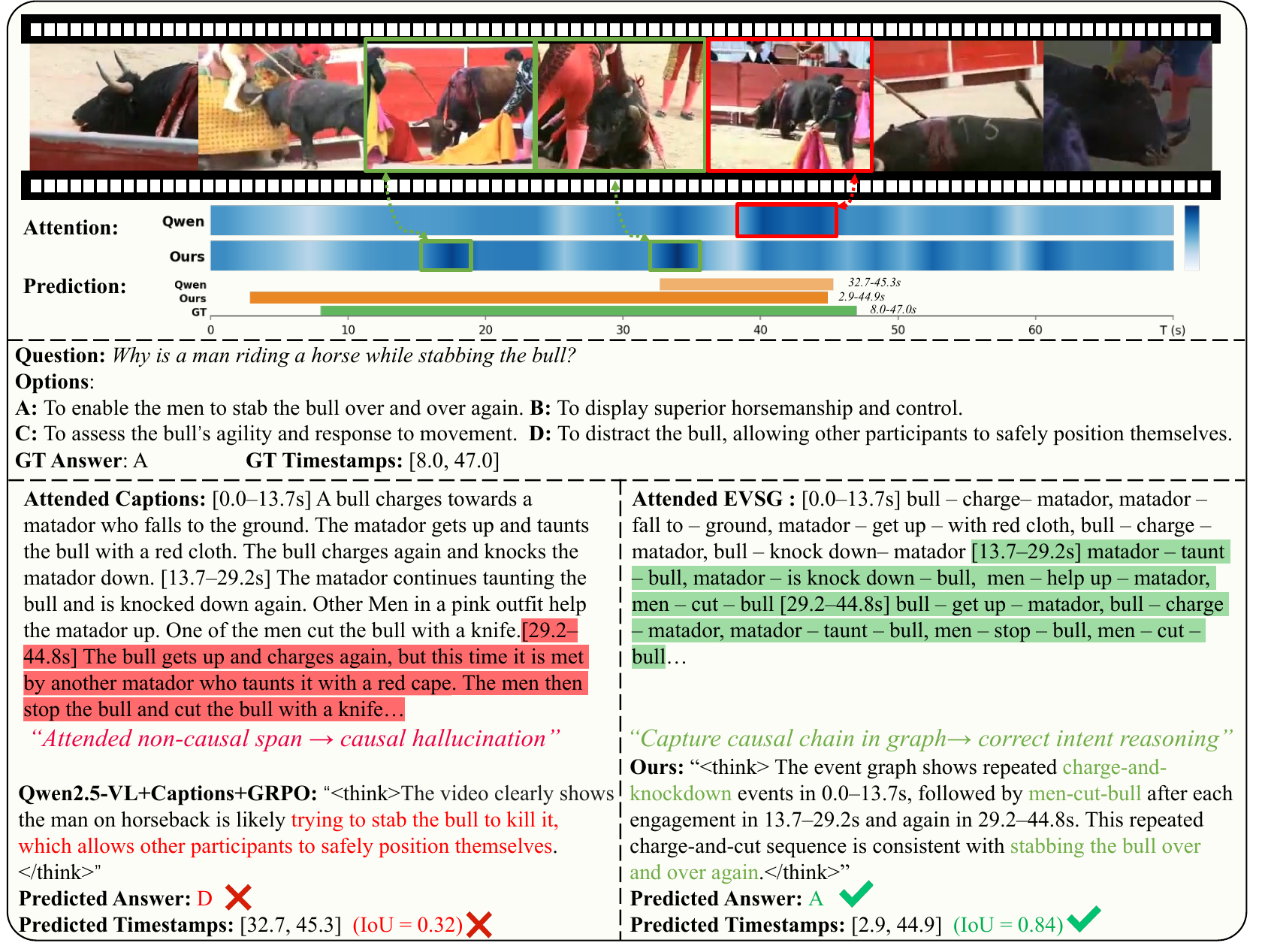}
  \vspace{-15pt}
  \caption{An illustration examples showing that \ourmeth{} improves causal reasoning in ReXTime~\cite{chen2024rextime} benchmark.
  }
  \label{fig:visualization2}
  \vspace{-10pt}
\end{figure}

\noindent\textbf{Improving causal reasoning.} 
Fig.~\ref{fig:visualization2} illustrates how EVSG guidance improves causal (means-ends) reasoning in the ReXTime benchmark~\cite{chen2024rextime}. Without EVSG, Qwen2.5-VL over-attends to a later span describing participants stopping and cutting the bull, forming a locally plausible but non-causal rationale and incorrectly predicting \textcolor{red}{Option D} (distraction to help others reposition). In contrast, \ourmeth{} leverages the EVSG to select and aggregate temporally linked subgraphs that recurrently encode the \emph{charge-knockdown} and \emph{taunt-cut} interactions across multiple cycles (e.g., 13.7-29.2s and 29.2-44.8s). This recurring structural pattern supplies the missing causal context, revealing that the horseback stabbing is part of a repeated charge-and-cut process rather than a one-off distraction manoeuvre. Consequently, \ourmeth{} correctly infers the causal intent as \textcolor{green}{Option A} and localises the corresponding interval with higher precision.

\begin{figure}[!h]
  \centering
  \includegraphics[width=1.0\linewidth]{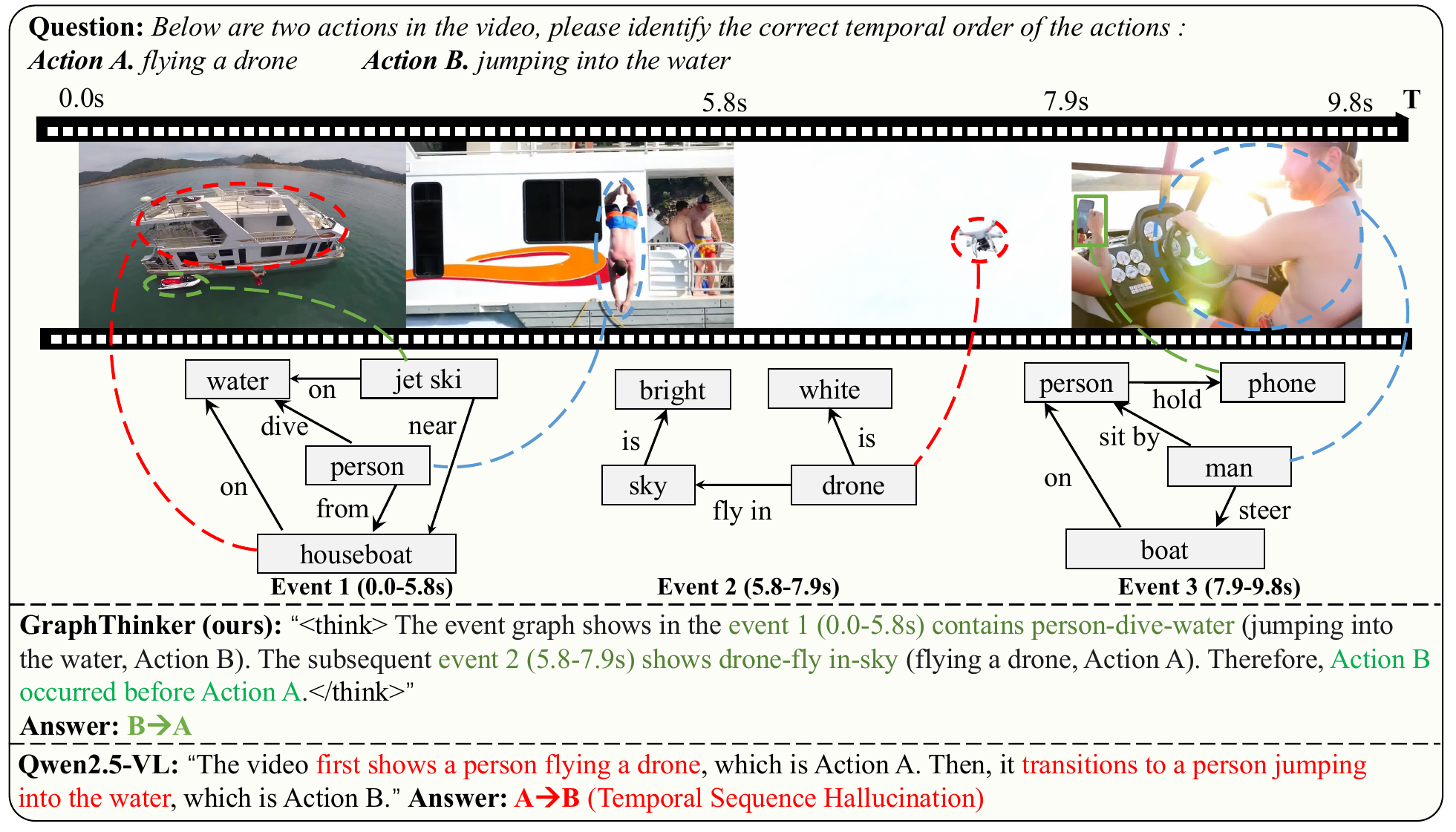}
  \vspace{-15pt}
  \caption{An illustration examples showing that \ourmeth{} mitigate Temporal Sequence Hallucination (TSH) in VidHalluc~\cite{li2025vidhalluc} benchmark.
  }
  \label{fig:hallucination1}
  \vspace{-10pt}
\end{figure}

\noindent\textbf{Mitigating temporal sequence hallucination.} Fig.~\ref{fig:hallucination1} demonstrates how \ourmeth{} mitigates temporal sequence hallucination in the VidHalluc benchmark~\cite{li2025vidhalluc} by enforcing event-level ordering through the EVSG. Since VidHalluc does not involve temporal grounding, temporal attention maps are not shown in these examples. The baseline (Qwen2.5-VL) relies on misinterpreted visual cues (e.g., the drone appearing prominently in later frames) and incorrectly narrates the video as ``fly a drone'' followed by ``jump into the water'' \textcolor{red}{(A$\rightarrow$B)}, despite the jump occurring earlier in the video. In contrast, \ourmeth{} constructs an EVSG that segments the video into time-grounded events and connects them with explicit temporal relations: Event~1 (0.0-5.8s) encodes the triplet \textit{person-dive-water} (Action~B), whilst the subsequent Event~2 (5.8-7.9s) encodes \textit{drone-fly in-sky} (Action~A). By reasoning over the temporal order of these subgraphs, \ourmeth{} correctly infers \textcolor{green}{B$\rightarrow$A}, preventing the model from temporally misaligned observations.

\begin{figure}[!h]
  \centering
  \includegraphics[width=1.0\linewidth]{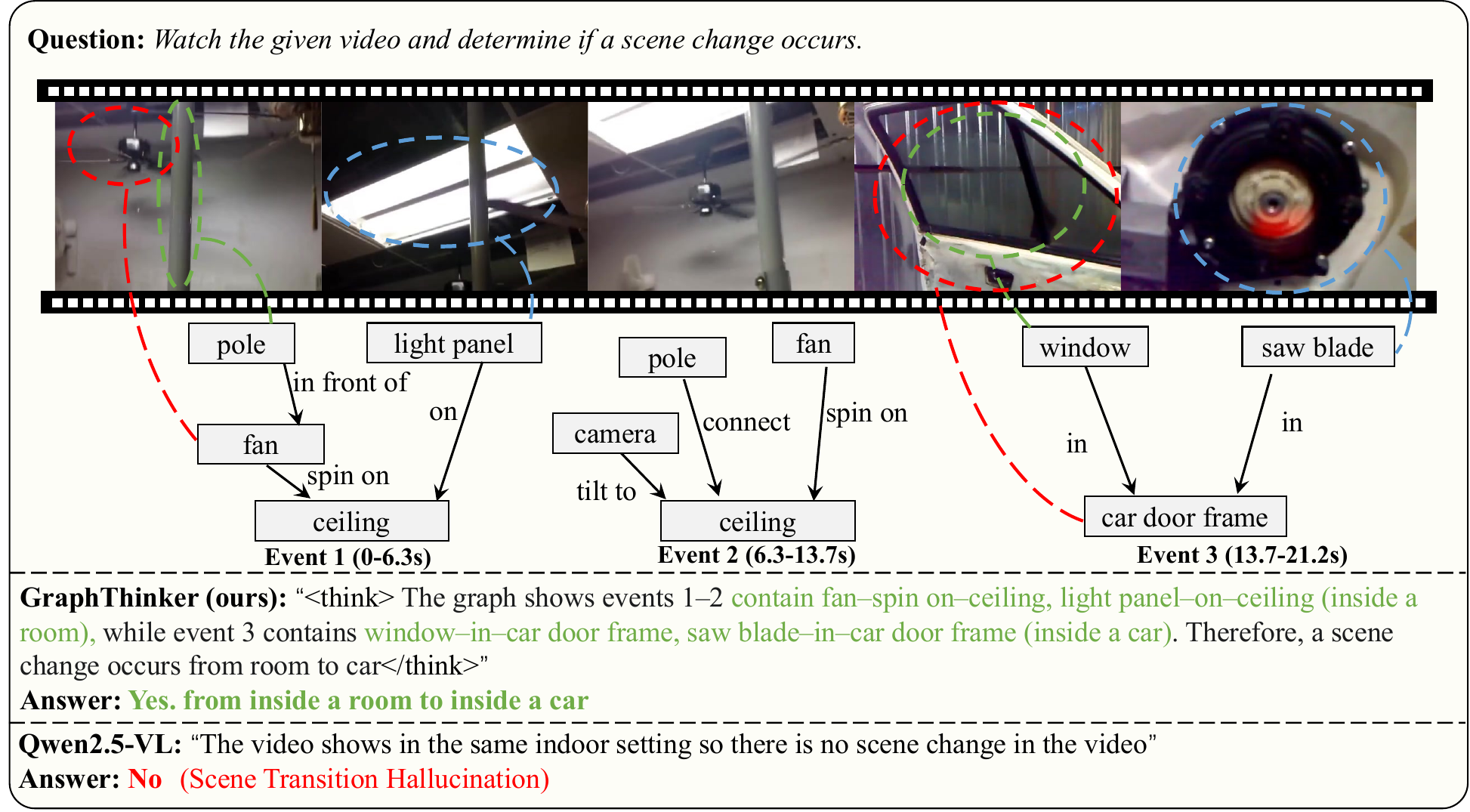}
  \vspace{-15pt}
  \caption{An illustration examples showing that \ourmeth{} mitigate scene transition hallucination (STH) in VidHalluc~\cite{li2025vidhalluc} benchmark.
  }
  \label{fig:hallucination2}
  \vspace{-10pt}
\end{figure}

\noindent\textbf{Mitigating scene transition hallucination.} Fig.~\ref{fig:hallucination2} presents a challenging scene transition hallucination case in the VidHalluc benchmark~\cite{li2025vidhalluc} in which the baseline is misled by apparent surface continuity. Although the early video clips repeatedly depict a ceiling fan and a light panel, Qwen2.5-VL treats the later shots as a mere viewpoint change and incorrectly predicts \textcolor{red}{\emph{No}} scene transition in the video. 
In contrast, \ourmeth{} uses the EVSG to compare event-level semantics across time: Events~1-2 share a consistent indoor-room context (\emph{fan--spin on--ceiling}, \emph{light panel-on-ceiling}), whereas Event~3 introduces a semantically distinct set of car-related entities and relations (\emph{window-in-car door frame}, \emph{saw blade-in-car door frame}). This explicit event partitioning and graph-based evidence selection renders the scene boundary salient, enabling \ourmeth{} to correctly detect a transition \textcolor{green}{from inside a room to inside a car}, mitigating scene transition hallucination.

\begin{figure}[!h]
  \centering
  \includegraphics[width=1.0\linewidth]{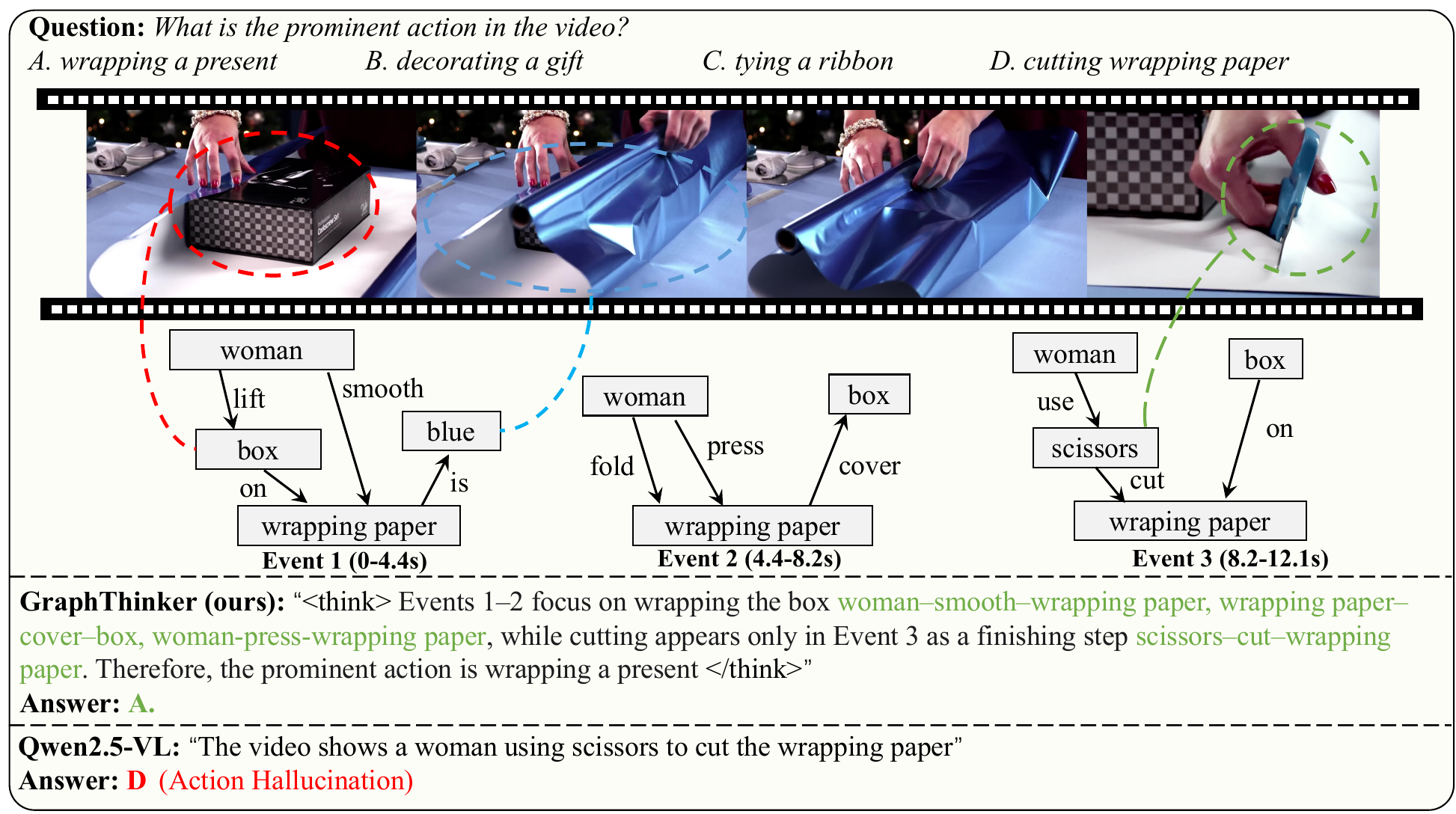}
  \vspace{-15pt}
  \caption{An illustration examples showing that \ourmeth{} mitigate action hallucination (ACH) in VidHalluc~\cite{li2025vidhalluc} benchmark.
  }
  \label{fig:hallucination3}
  \vspace{-10pt}
\end{figure}

\noindent\textbf{Mitigating action hallucination.}
Fig.~\ref{fig:hallucination3} illustrates an action hallucination case in the VidHalluc benchmark~\cite{li2025vidhalluc} arising from over-reliance on a salient sub-action. Qwen2.5-VL is distracted by the final cutting moment encoded in Event~3 (\emph{scissor-cut-wrapping paper}) and incorrectly predicts \textcolor{red}{Option~D: cutting wrapping paper} as the dominant action. In contrast, \ourmeth{} aggregates evidence across all events via the EVSG: Events~1-2 capture the primary wrapping procedure through the triplets \emph{woman-smooth-wrapping paper}, \emph{wrapping paper-cover-box}, and \emph{woman-press-wrapping paper}, whilst cutting appears only as a brief finishing step in Event~3. This event-level evidence aggregation enables \ourmeth{} to predict the correct dominant action, \textcolor{green}{Option~A: wrapping a present}, effectively mitigating action hallucination caused by bias towards visually salient sub-actions.

\clearpage

\end{document}